%% file: main.tex
\documentclass[11pt]{article}
\input{preamble.tex}

\title{Disentangling Statistical Preemption from Entrenchment in Language Models' Avoidance of Overgeneralization}

\author{Yixuan Wang \\
  University of Waterloo \\
  Vector Institute \\
  \texttt{tyxw@uwaterloo.ca} \\\And
  Freda Shi \\
  University of Waterloo \\
  Vector Institute \\
  Canada CIFAR AI Chair \\
  \texttt{fhs@uwaterloo.ca} \\\And
  Kanishka Misra \\
  Department of Linguistics \\
  University of Texas at Austin \\
  \texttt{kmisra@utexas.edu} \\}

\begin{document}

\maketitle

\begin{abstract}

How do learners avoid overgeneralizations such as \textit{Tom laughed me} without explicit negative evidence? Constructivists have posited two proposals that describe indirect negative evidence against overgeneralizations: preemption (which privileges exposure to near-synonymous construction---e.g., \textit{she made him laugh}) vs. entrenchment (all exposures to a verb's grammatical usages, including cases like \textit{He laughed}). We disentangle these hypotheses by running controlled rearing experiments on LMs trained on child-caregiver conversations, where we systematically remove preemptive vs. non-preemptive evidence. We find that while LMs avoid overgeneralizations, they do not show preemption at a verb-specific level, instead showing evidence of abstract preemption. Combined with results from analyzing the LMs' training dynamics, we find that LMs treat competing structures as indirect positive---as opposed to negative---evidence in the verb-specific condition. Insofar as preemption is the more plausible route to avoiding overgeneralizations in humans, our results suggest the need for there to be greater sensitivity to indirect negative evidence in neural network learners, and motivate new human experiments to test preemptive effects of verbs beyond the target verb.
\end{abstract}


\input{chapters/01-intro.tex}

\input{chapters/02-prelim.tex}
\input{chapters/03-methods.tex}
\input{chapters/04-experiment-precondition.tex}
\input{chapters/05-experiment-preemption.tex}
\input{chapters/06-experiment-moment.tex}

\input{chapters/99-closing.tex}

\bibliography{ref}

\appendix

\input{chapters/a1-technical}

\input{chapters/a2-additional-results.tex}
\input{chapters/a3-guo}

\input{chapters/a4-alt}

\end{document}

%% file: preamble.tex
\usepackage[final]{acl}

\usepackage{times}
\usepackage{latexsym}
\usepackage[T1]{fontenc}
\usepackage[utf8]{inputenc}
\usepackage{microtype}
\usepackage{inconsolata}

\usepackage{xcolor}
\usepackage{import}

\usepackage{amsmath}
\usepackage{amssymb}
\usepackage{bbm}
\usepackage{mathtools}
\usepackage[normalem]{ulem}

\usepackage{graphicx}
\graphicspath{{./assets/figures}}
\usepackage{subcaption}
\usepackage{multirow}
\usepackage{array}
\usepackage{tcolorbox}
\usepackage{enumitem}
\usepackage{fvextra}

\usepackage{booktabs}
\usepackage{langsci-gb4e}

\usepackage{hyperref}

\input{utils/package-cleveref.tex}

\input{utils/package-todonotes.tex}

\input{utils/package-langsci-gb4e.tex}

\input{utils/macros.tex}

\input{utils/shorthands.tex}

%% file: utils/package-cleveref.tex
\usepackage[capitalize,noabbrev]{cleveref}

\crefname{chapter}{Chapter}{Chapters}
\crefname{section}{\S}{\S\S}
\Crefname{section}{\S}{\S\S}
\crefname{table}{Table}{Tables}
\crefname{figure}{Figure}{Figures}
\crefname{algorithm}{Algorithm}{}
\crefname{equation}{Eq.}{}
\crefname{appendix}{Appendix}{}
\crefformat{section}{\S#2#1#3}
\crefname{lemma}{Lemma}{Lemmas}
\crefname{proposition}{Proposition}{Propositions}
\crefname{definition}{Definition}{Definitions}
\crefname{corollary}{Corollary}{Corollaries}
\crefname{example}{Example}{Examples}
\crefname{problem}{Problem}{Problems}
\crefname{remark}{Remark}{Remarks}

\crefname{ExNo}{example}{examples}
\Crefname{ExNo}{Example}{Examples}
\crefname{SubExNo}{example}{examples}
\Crefname{SubExNo}{Example}{Examples}
\crefname{SubSubExNo}{example}{examples}
\Crefname{SubSubExNo}{Example}{Examples}
\crefformat{ExNo}{(#2#1#3)}
\crefformat{SubExNo}{(#2#1#3)}
\crefformat{SubSubExNo}{(#2#1#3)}
\crefrangeformat{ExNo}{\mbox{(#3#1#4--#5#2#6)}}
\crefrangeformat{SubExNo}{\mbox{(#3#1#4--#5#2#6)}}
\crefrangeformat{SubSubExNo}{\mbox{(#3#1#4--#5#2#6)}}
\crefmultiformat{ExNo}{(#2#1#3}{, #2#1#3)}{, #2#1#3}{, #2#1#3)}
\crefmultiformat{SubExNo}{(#2#1#3}{, #2#1#3)}{, #2#1#3}{, #2#1#3)}
\crefmultiformat{SubSubExNo}{(#2#1#3}{, #2#1#3)}{, #2#1#3}{, #2#1#3)}
\crefrangemultiformat{ExNo}{(#3#1#4--#5#2#6}{, #3#1#4--#5#2#6)}{, #3#1#4--#5#2#6}{, #3#1#4--#5#2#6)}
\crefrangemultiformat{SubExNo}{(#3#1#4--#5#2#6}{, #3#1#4--#5#2#6)}{, #3#1#4--#5#2#6}{, #3#1#4--#5#2#6)}
\crefrangemultiformat{SubSubExNo}{(#3#1#4--#5#2#6}{, #3#1#4--#5#2#6)}{, #3#1#4--#5#2#6}{, #3#1#4--#5#2#6)}

%% file: utils/package-todonotes.tex
\usepackage[
    textsize=tiny,
    colorinlistoftodos
]{todonotes}

\newcommand{\note}[4][]{{\todo[caption={},#1]{\textcolor{#3}{\textbf{#2}}~#4}}}

\todostyle{orange}{color=orange!10,bordercolor=orange!90,linecolor=orange!90}
\newcommand{\yixuan}[2][]{{\note[orange,#1]{yixuan}{orange!90}{#2}}}
\newcommand{\Yixuan}[2][]{{\note[orange,inline,#1]{yixuan}{orange!90}{#2}}}

\definecolor{kcyan}{RGB}{27, 108, 168}
\todostyle{kcyan}{color=kcyan!10,bordercolor=kcyan!90,linecolor=kcyan!90}

\definecolor{tticblue}{RGB}{0, 94, 184}
\todostyle{tticblue}{color=tticblue!10,bordercolor=tticblue!90,linecolor=tticblue!90}

%% file: utils/macros.tex
\makeatletter
\ifacl@anonymize
  \newcommand{\blind}[1]{}
\else
  \newcommand{\blind}[1]{#1}
\fi
\makeatother

%% file: utils/shorthands.tex
\newcommand{\ogened}{overgeneralized}
\newcommand{\ogenzation}{overgeneralization}
\newcommand{\Ogenzation}{Overgeneralization}

\newcommand{\variantRemoveP}{\textsc{remove-p}}
\newcommand{\variantKeepP}{\textsc{keep-p}}
\newcommand{\removep}{\textsc{remove-p}}
\newcommand{\keepp}{\textsc{keep-p}}

%% file: chapters/01-intro.tex

\section{Introduction}
\label{sec:intro}

One of the hallmarks of human cognition is our ability to generalize productively. For instance, on hearing \cref{ex:boil-intransitive}, competent speakers of English use their experience of intransitive and transitive verbs (\textit{break, roll}) to effortlessly produce \cref{ex:boil-causative}. But their generalization of this pattern is also constrained---e.g., while we are perfectly fine with \cref{ex:laugh-intransitive}, we do not produce \cref{ex:laugh-causative}.



\begin{exe}
  \ex \label{ex:boil} \begin{xlist}
    \ex[]{The water is boiling.} \label{ex:boil-intransitive}
    \ex[]{Dad boiled some water.} \label{ex:boil-causative}
  \end{xlist}

  \ex \label{ex:laugh} \begin{xlist}
    \ex[]{The baby laughed.} \label{ex:laugh-intransitive}
    \ex[*]{Mom laughed the baby.} \label{ex:laugh-causative}
  \end{xlist}
\end{exe}

\begin{figure}[!t]
    \centering
    \includegraphics[width=0.8\columnwidth]{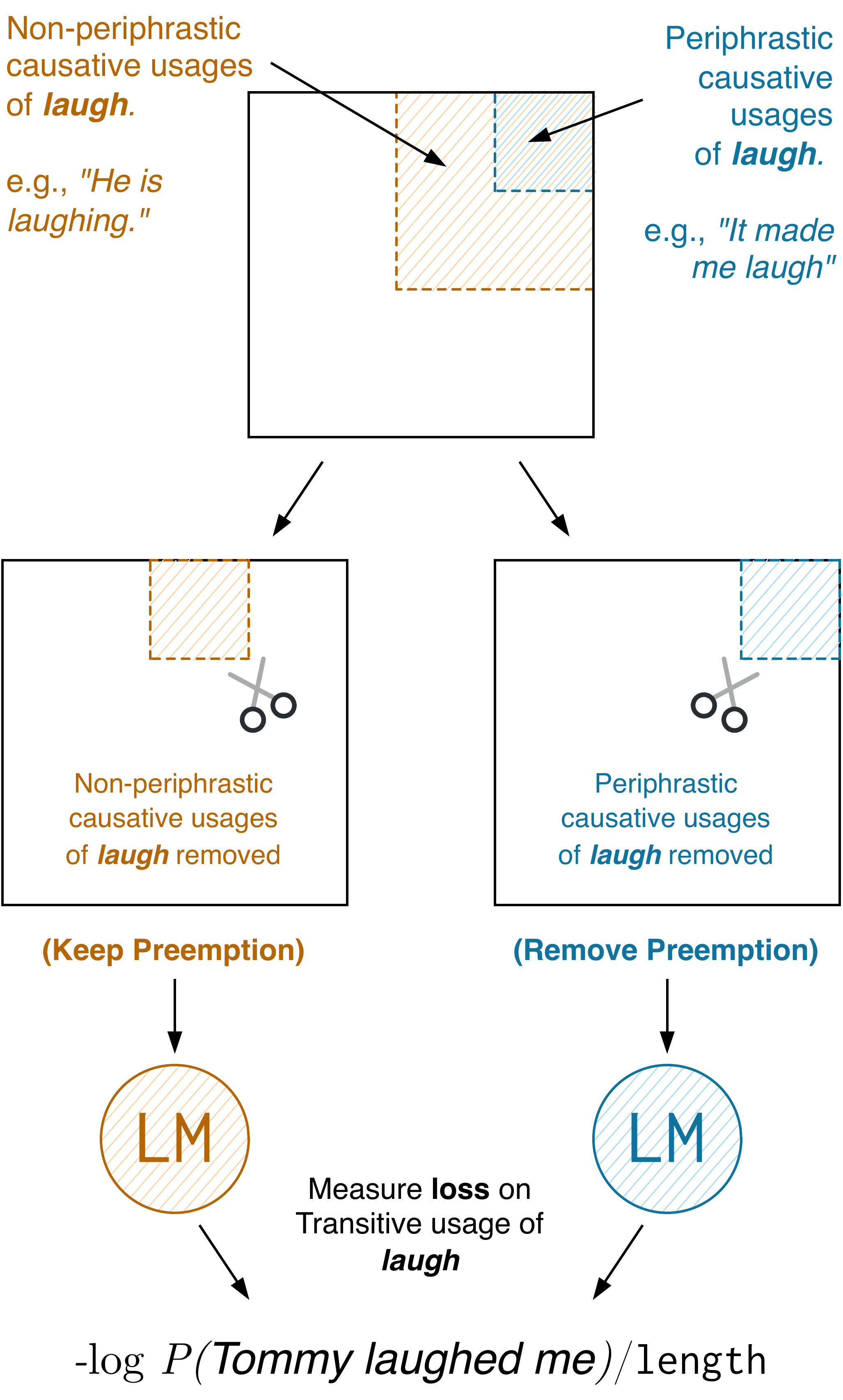}
    \caption{Overview of our experimental paradigm. For a given verb (\textit{laugh}), we train LMs on two versions of our corpus, created by removing: 1) the preemptive evidence for that verb; and 2) same amount of non-preemptive evidence. We then compare the behavior of these LMs to disentangle preemption from entrenchment in explaining that verb's overgeneralization.}
    \label{fig:pipeline-description}
\end{figure}

Children often tend to \textit{overgeneralize} the above pattern by indeed producing sentences like \cref{ex:laugh-causative} \citep[][i.a.]{bowerman1988no, braine1995verb, akhtar1997young, brooks1999children}, but eventually \textit{retreat} and constrain their generalizations. They do so \textit{without} much explicit negative evidence \citep{brown1970derivational, baker1979syntactic}---caregivers rarely ever correct children's grammatical errors systematically \citep{chouinard2003adult}. How, then, are such constraints and restrictions learned?

In the realm of constructivist and usage-based approaches to language acquisition \citep{goldberg1995constructions, tomasello2003constructing, goldberg2005constructions, rowland2013understanding, rowland2025constructing}, there are two closely related proposals that attempt to explain this avoidance of overgeneralizations: \textit{preemption} and \textit{entrenchment}. 
Preemption suggests that given two near-synonymous constructions $A$ and $B$, exposure to a verb used in $A$ in contexts where the verb could have been used in either construction serves as evidence \textit{against} the verb's usage in $B$ \citep{goldberg1995constructions, brooks2002does, boyd2011learning}. In the case of \textit{laugh}, preemption predicts that the transitive \textit{*Mom laughed the baby} is \textit{preempted} by the periphrastic causative usage \textit{Mom made the baby laugh}, because in situations where \textit{X} did something that caused \textit{Y} to \textit{laugh}, the utterance \textit{X made Y laugh} was used instead of \textit{X laughed Y}.
Entrenchment, on the other hand, does not privilege near-synonymous constructions or specific contexts \citep{theakston2004role, stefanowitsch2008negative, ambridge2008effect}. Here, the usage of the verb in \textit{any} attested construction is evidence against the unattested usage---entrenchment not only considers periphrastic causative usages of \textit{laugh} in blocking its transitive usage, but also other constructions like the intransitive (as in \cref{ex:laugh-intransitive}). These two proposals have been up for debate for about 40 years \citep{rowland2013understanding}.

There are more similarities between these proposals than there are differences: 1) both explanations rely on \textit{indirect} negative evidence, since they posit \textit{other} constructions as evidence against overgeneralization; 2) both are statistical in nature, since they rely on the frequency of exposure to the indirect evidence---more frequent this exposure is, stronger is the blocking effect; and most importantly, 3) entrenchment is a strict superset of preemption, since the evidence that falls under the purview of preemption (near-synonymous construction) is always part of the evidence considered by entrenchment. 
The latter point explains why attempts to disentangle the two proposals in the laboratory \citep{brooks2002does, ambridge2015preemption, ambridge-etal-2018-effects, bidgoodVerbArgumentStructure2021} often suffer from multi-collinearity issues in both variables. In these studies, participants' acceptability judgments on unconventional forms are predicted using the frequencies of the preemptive usages of a verb (preemption) as well as that of the total frequency of the verb (entrenchment) sourced from openly available corpora, rather than the learners' own experience (which is unobtainable for humans). Therefore, apart from multicollinearity (due to the overlap in preemptive and entrenching evidence), these experiments and analysis are also unable to shed light on the causal status of these proposals.

To adjudicate between these two proposals, we conduct ``controlled rearing'' \citep{frank2023bridging, misra-mahowald-2024-language} experiments on Language Models (LMs), where we systematically manipulate the learning experience of LMs to disentangle between preemption and entrenchment. As a case study, we apply our methods to explaining transitive overgeneralizations of (usually) intransitive verbs such as \textit{laugh, smile, cry}, etc., as this is among the quintessential examples of entrenchment and preemption analyses over the past 30 years \citep{akhtar1997young, ambridge2015preemption, bidgoodVerbArgumentStructure2021}. While analyses on LMs cannot \textit{directly} be taken to have implications for humans \citep[cf.][]{warstadt2022artificial, portelance2024roles, misra2024generating, futrell2025linguistics}, such experiments allow us to conclude about the viability of the two proposals and how (and whether) they might arise from the general purpose statistical learning of the kind employed by LMs. Controlled rearing allows us to do this by considering counterfactual learning scenarios---e.g., ``what if there was no preemptive evidence in the learner's training data?'' Evidence of a privileged role of
preemptive evidence would suggest that LMs acquire restraints on their generalization behavior from fine-grained sensitivities to meaning, since preemption privileges contexts where either construction could have been used to communicate the same intent or semantics.
More broadly, this allows us to understand if similar constraints arise within LMs trained on cognitively plausible amounts of language-only exposure, or if explicit evidence of semantics is needed \citep{abend2017bootstrapping, yedetore-kim-2024-semantic}.


Our methods and research questions also make methodological contributions to the burgeoning program of ``controlled rearing'' \citep{frank2023bridging, misra-mahowald-2024-language} or ``filtered corpus training'' \citep{patil2024filtered}, which has established itself as an important paradigm in understanding the role of input in shaping a statistical learner's generalizations. Most applications of this method have primarily demonstrated how LMs generalize to ``novel'' usages in the absence of direct or impoverished evidence \citep{jumelet-etal-2021-language, misra-mahowald-2024-language, patil2024filtered, yao2025both, yang2026unified}. That is, they only focus on indirect \textit{positive} evidence, since they investigate what facilitates generalization, and not what \textit{blocks} it. An exception to this is work by \citet{leong2026manipulating}, who focus on explicitly understanding signals that \textit{block} models' usage of a structure, but focus on a different construction (active/passive), and do not consider the difference between preempting and non-preempting evidence (i.e., they consider entrenchment as a whole). This positions our work uniquely within this literature, since we aim to causally disentangle preemption and entrenchment, and explicitly measure their viability as sources of indirect \textit{negative} evidence.


\subsection{Summary of Findings}

Before testing what explains models' avoidance of overgeneralization, we first measure if they learn the appropriate generalizations in the first place---i.e., finding transitive usages of intransitive verbs less probable than their periphrastic causative counterparts. Finding that they do (\cref{sec:precondition}) for six different verbs, we then run our controlled rearing studies to disentangle preemption from entrenchment (\cref{sec:preemption}). Here, we test if preemptive usages of a given verb (i.e. their periphrastic causative usages like in \textit{Tom made me laugh}) have a privileged role over non-preemptive ones (e.g., \textit{I laughed.}), by training LMs on corpora where each of them are removed in equal amounts separately. 
Using LMs' negative log-likelihood loss (NLL) on the unconventional, unseen transitive usages of these verbs as our measure of overgeneralization, we considered two different manipulation types: 1) \textbf{verb-specific}, where for a verb only its own usages were manipulated; and 2) \textbf{abstract}, where usages of all verbs were manipulated, to quantify if the evidence of preemption comes from higher order sources.

We do not find clear evidence for preemption in the verb-specific condition---LMs' NLL of the unconventional transitive usage was greater when preemptive exposures were removed than when they were kept in. By contrast, preemption would predict that in the absence of preemptive exposures, models should find transitive usages \textit{more} likely. The abstract condition, however, did show relatively larger effects of preemption, though in absolute terms these were generally weak. This finding is consistent with the hypothesis that models treat the periphrastic causative construction as indirect positive evidence, common in other controlled rearing studies. That is, the models might have inferred the relation between transitive and periphrastic causatives via other verbs for which both are acceptable (e.g., \textit{She boiled the water} vs. \textit{She got the water to boil}), which explains why this is not observed when all periphrastic causatives are removed (in the abstract condition).

To further explore this finding of positive---as opposed to negative---evidence, we investigated the change in the model's behavior on unseen transitive and periphrastic causative sentences before and after a periphrastic causative usage was encountered by the model during training. We find the model's generalization behavior on both constructions to be highly correlated, further supporting the indirect positive evidence hypothesis---the presence of periphrastic causatives is tightly associated with the likelihood of transitive usages.

Insofar as preemption occurs and drives retreat from overgeneralization in humans \citep{samara2025learners}, our verb-specific results indicate the need for LM learners to model fine-grained indirect negative evidence. We speculate that this could come from more explicit semantic signals about the context in which either construction is used, or from morphological constraints, where the concept of preemption originates from \citep{kiparsky1973elsewhere, aronoff1976word}. 
At the same time, our finding of abstract preemption motivates new experiments that test this finding in humans via an artificial language learning design, since preemption for argument structure alternations is often treated as a verb-specific effect \citep{goldberg2005constructions, ambridge2015preemption, ambridge-etal-2018-effects, bidgoodVerbArgumentStructure2021}. Our code and data are available at \url{https://github.com/Yixuan-Wang/verb-overgen}.

%% file: chapters/03-methods.tex
\section{Data, Model, and Methods}
\label{sec:method}


\subsection{Training Data}
\label{sec:training-data}

To remain as close as possible to the input and constraints available to children, we use the English subset of the Child Language Data Exchange System (CHILDES) dataset \citep{macwhinney2000childes} as the training dataset for our models. CHILDES consists of human-transcribed conversations between children and caregivers. The corpus consists of approximately 5M utterances, amounting to a total of 26M space-delimited words. 
Importantly, CHILDES also comes with morphological and syntactic annotations.\footnote{\url{https://talkbank.org/0info/manuals/CHAT.pdf}} This allows us to precisely detect phenomena we are interested in, avoiding any potential errors from existing automatic tools \citep[cf.][]{yang-etal-2025-ud, padovani2026cait}.

As a preprocessing step, we lemmatize the corpus and strip away morphological inflections. This allows models to represent tense-inflected versions of verbs using the same token (\textit{laughed} $\rightarrow$ \textit{laugh}, \textit{laughing} $\rightarrow$ \textit{laugh}), thereby giving them the maximum chance---if any---to use inflected forms of the same verb as collective evidence.\footnote{We show results with no lemmatization in \cref{sec:appendix-bpe}.} 
All subsequent experimental data, including evaluation, use the lemmatized sentences.



\subsection{Model}
\label{sec:model}

We use a word-level tokenizer for the lemmatized training data, with a vocabulary size of 37,482.
Our learners are autoregressive LMs with the GPT-2 architecture \citep{radford-etal-2019-gpt2}.
Each LM has 12 layers and 12 attention heads per layer, with a hidden size of 768, resulting in a total of 115M parameters. 
Each model is trained for 3 epochs, and training is repeated with 5 random seeds to marginalize nondeterminism, including parameter initialization and training data shuffling.

\subsection{Disentangling Preemption from Entrenchment}
\label{sec:disentangle}

Because preemption is necessarily a subset of entrenchment (see \Cref{sec:intro}), disentangling the two boils down to testing whether the hypothesized preemptive evidence has a \textit{privileged} role compared to those not considered as part of preemption. This effectively partitions the set of usages of a particular verb to preemptive and non-preemptive. 
In the context of transitive overgeneralizations (\textit{she laughed me}), the preemptive evidence is the periphrastic causative construction (\textit{she made me laugh}), and the non-preemptive ones are all other types of usages (\textit{he is laughing, she laughed at me,} etc.).
This give us two kinds of manipulations for our controlled rearing experiments: 1) \textbf{\removep{}}, where we remove \textit{all} occurrences of the periphrastic causative construction usages for a given verb (or a set of verbs); and 2) \textbf{\keepp{}} (or \textsc{remove-non-p}), where we remove the same number of occurrences of a given verb (or set of verbs) as in \removep{}, that \textit{do not} use it in the periphrastic causative construction. 

The raw frequency of a verb of interest in both cases is reduced by the same amount, but the structures being manipulated are different. Insofar as models assign a special role to preemptive evidence, we should expect them to retreat further from overgeneralizations in \keepp{}, which retains the preemptive evidence, than in \removep{}. In other words, the model should find transitive overgeneralizations more likely under \removep{} than under \keepp{}, since the hypothesized critical evidence against transitive usages is absent in \removep{}.

In our experiments, we primarily focus on six verbs that have been described to be `intransitive-only' by \citet{levin-1993-english}: \textit{go, laugh, cry, sleep, sneeze,} and \textit{smile}. Using them transitively almost always constitutes an overgeneralization.\footnote{There can be creative, usages of these verbs that appear transitive---e.g., in the \textsc{caused-motion} construction: \textit{She sneezed the foam off the cappuccino} \citep{goldberg2005constructions}, but these do not have the same meaning as the causative context---the agent (she) is not making the object (foam) sneeze.} \Cref{tab:verb-counts} shows the usage statistics of these verbs in CHILDES, both as a verb in general, and also more specifically in the periphrastic causative construction.
The periphrastic usages of these verbs are detected using the gold-dependency parses provided in the CHILDES dataset.
A periphrastic causative construction is identified if the matrix verb (e.g., \textit{make}, \textit{let}, etc.) has a complement child (\texttt{COMP}) and a patient in the dependency parse.
Technical details about the identification of periphrastic causative constructions can be found in \cref{sec:causative-identification}.

\begin{table}[!t]
\centering
\resizebox{0.65\columnwidth}{!}{%
\input{assets/tables/train_verb_count.tex}
}
\caption{Usage statistics of the verbs of interest in the training set, including the number of times each word appears in a Periphrastic Causative (PC) construction, and the total number of times the word is used as a verb.}
\label{tab:verb-counts}
\end{table}

\subsection{Measuring Overgeneralization Effects}
\label{sec:measure}

\paragraph{Evaluation Data}
Our evaluation data constitutes a pair of transitive and periphrastic causative sentences containing the abovementioned verbs and a set of nouns that occur in the CHILDES dataset. We construct lemmatized versions of these sentences by using the template ``\texttt{S V O}'' (e.g., \textit{Tommy laugh Mary}) for the transitive structure, and ``\texttt{S make O V}'' (e.g., \textit{Tommy make Mary laugh}) for the periphrastic causative structure. Here, \texttt{S}, \texttt{O}, and \texttt{V} are the subject, object, and the verb lemma, respectively, such that none of the sentences occur in CHILDES. In total, we generated 100 pairs for each verb. We denote transitive sentences as $s^-$, and periphrastic causative ones as $s^+$. 

\paragraph{Measure}
We measure the overgeneralization effect for a given verb as the average negative log-likelihood loss (NLL) of the transitive sentences for that verb in our evaluation data. Specifically, given an LM $P^{\mathcal{D}}$, trained on a corpus $\mathcal{D}$, and a transitive sentence, $s^- = x^-_1 \ldots x^-_{|s^-|}$, we measure:
\begin{align*}
    \ell(s^-; P^{\mathcal{D}}) = - \frac{1}{|s^-|} \sum_{i=1}^{|s^-|} \log P^{\mathcal{D}}\left(x^{-}_{i}\mid x^{-}_{<i}\right)
\end{align*}
This allows us to compare models trained on corpora with different amounts of indirect negative evidence in terms of how (un)likely or surprising they find transitive usages of a verb. If the evidence contained in a corpus $\mathcal{D}_1$ better suppresses the overgeneralization of a verb than does the evidence in $\mathcal{D}_2$, then we expect to observe a higher NLL on $P^{\mathcal{D}_1}$ than on $P^{\mathcal{D}_2}$, i.e., $\ell(s^-; P^{\mathcal{D}_1}) > \ell(s^-; P^{\mathcal{D}_2})$. 

It might be tempting to treat overgeneralization effect as the \textit{preference} of the periphrastic causative evaluation sentence $s^+$ over its overgeneralized transitive form $s^-$, by measuring the difference between their NLLs. Afterall, these two constructions are hypothesized to compete \citep{goldberg2005constructions}. However, we argue that this metric does not allow us to explain overgeneralization effects, which are explicitly concerned with what allows the learner to conclude that the unattested form (here, transitive) is unacceptable. This is because our experiments (as we will show in \cref{subsec:preemption-manipulations}) directly manipulate the presence of the periphrastic causative exposures available to the model. Therefore, changes to this alternate measure could be completely driven by the likelihood of the conventional, periphrastic causative sentence (whose exposure has been manipulated) rather than the unlikelihood of the transitive counterpart. Indeed, such a measure has been used in previous work \citep{guo2026do}, but due to the above reason, it can paint an illusory picture of preemption (see more in \cref{sec:rebuttal-guo} and \cref{sec:guo-results} where we compare our results to the one had we used this alternate measure).

%% file: assets/tables/train_verb_count.tex
\begin{tabular}{lrr}
\toprule
\textbf{Verb} & \textbf{PC Usage} & \textbf{Verb Usage} \\
\midrule
\textit{go} & 1295 & 316515 \\
\textit{laugh} & 99 & 1781 \\
\textit{cry} & 49 & 6166 \\
\textit{sleep} & 33 & 8614 \\
\textit{sneeze} & 31 & 461 \\
\textit{smile} & 15 & 724 \\
\bottomrule
\end{tabular}

%% file: chapters/04-experiment-precondition.tex
\section{Preconditions to Not Overgeneralizing}
\label{sec:precondition}

Before analyzing how indirect negative evidence might facilitate our LMs to avoid overgeneralization, we first establish if the model trained on the full CHILDES corpus learns the appropriate generalization behavior---that the transitive usages of our target verbs is indeed unacceptable, relative to their periphrastic causative usages. This is the minimal precondition the model must satisfy before we can use it to conclude about the factors that affect its generalization, following \citet{misra2024generating}. 

\begin{figure}[!t]
    \centering
    \includegraphics[width=\columnwidth]{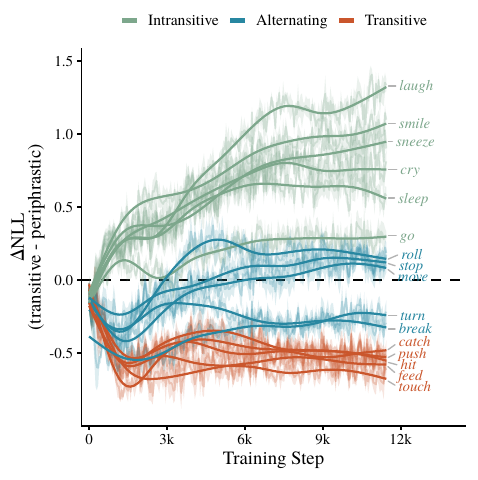}
    \caption{Difference in NLL of transitive and periphrastic causative sentence pairs for intransitive (target), alternating, and transitive verbs at every 50 steps during training. Smoothed curves are estimated using a generalized additive model \citep{hastie1986generalized}. Across all intransitive verbs, models show greater loss on transitive sentences than on their periphrastic causative counterparts, suggesting that they have acquired the right preference for these verbs. This is in contrast to alternating and transitive verbs, for which models show relatively lower $\Delta$NLL values, as expected.}
    \label{fig:precondition-results}
\end{figure}

To test this criterion, we compare the NLL of the transitive evaluation sentences (i.e., the unconventional forms, $s^-$) to their periphrastic causative counterparts ($s^+$). Insofar as the model has learned the right generalizations, we expect this measure to be a positive value, since the loss on the transitive should be greater than that on the periphrastic causative, since the latter is more acceptable. In addition, these results should not just be explained away by the fact that the model has a periphrastic causative bias. That is, its preference for verbs that do not permit the periphrastic causative usage (e.g., \textit{hit}) and are instead transitive-biased should show a reverse effect. To account for this additional precondition, we compare the $\Delta$NLL values for intransitive verbs to 5 alternating verbs (\textit{break}, \textit{move}, \textit{stop}, \textit{turn}, \textit{roll}) and 5 verbs that have a transitive bias (\textit{catch}, \textit{feed}, \textit{hit}, \textit{push}, \textit{touch}), all based on \citet{levin-1993-english}'s classification. We expect the transitive-biased verbs to show the opposite behavior as our target intransitive verbs (i.e., negative $\Delta$NLL values), and alternating verbs to be somewhere in between. We create 100 pairs of sentences for each of these verbs using the same procedure as our target verbs (see \Cref{sec:disentangle}). We measure $\Delta$NLLs for sentence pairs of all three classes of verbs (Intransitive, Alternating, and Transitive) across model training steps (every 50 batches), and show the resulting curves in \Cref{fig:precondition-results}.\footnote{A cleaner version with just the final checkpoints can be found in \Cref{app:final-ckpt-exp1}.}

We see from this figure that our models show clear evidence of expected behavior, as described above. Our intransitive verbs show generally positive $\Delta$NLLs, while the transitive verbs show generally negative $\Delta$NLLs, and alternating verbs are squarely in the middle. Overall, these results suggest that our LMs are well-positioned to answer questions about the factors that lower or increase (un)likelihood on the unconventional form for our target intransitive verbs. 

%% file: chapters/05-experiment-preemption.tex
\section{Verb-specific and Abstract Preemption}
\label{sec:preemption}

Having shown that the models indeed learn the appropriate generalization, we now turn to disentangling preemption from entrenchment. We compare the role of preemptive and non-preemptive evidence in the model's behavior on the unconventional, overgeneralized, transitive usages of our target verbs. That is, we compare models trained on manipulations specified by \keepp{} vs. \removep{}.

\subsection{Manipulations}
\label{subsec:preemption-manipulations}

We consider different levels at which preemption (or non-preemption) might affect model behavior: 
\paragraph{Verb-specific} This condition specifies the canonical form of preemption \citep{goldberg2005constructions, ambridge2015preemption, ambridge-etal-2018-effects, bidgoodVerbArgumentStructure2021}. Here, preemption for a given verb is only measured with respect to the exposure of the learner to the preemptive evidence for that particular verb, regardless of the learner's experience with usages of other verbs. Therefore, \keepp{} and \removep{} in this condition are only concerned with the periphrastic causative usages of a given verb at a time. This results in a total of 60 manipulations, resulting in 5 random seeds, 2 manipulation types, and 6 verbs.

\paragraph{Abstract} 
The abstract condition considers the effect of other verbs' preemptive usages in the learner's experience. Here, we measure the effect of the whole category that is denoted by the structure specified as the preemptive evidence. That is, for \removep{}, we remove periphrastic causative usages of \textit{all} verbs ($N$=7,452), and for \keepp{}, we remove equivalent amounts of (randomly sampled) non-periphrastic causative usages of all verbs. This results in a total of 10 manipulations, resulting from 5 seeds and 2 conditions.

\subsection{Comparisons to Alternating Verbs}
\label{sec:breakstop}
We compare our controlled rearing results on the six target verbs to two additional verbs---namely, \textit{break} and \textit{stop}. These verbs permit both transitive and intransitive usages, in addition to periphrastic causative ones---namely, \textit{break} and \textit{stop}. This results in 20 more verb-specific manipulations (2 verbs, 5 seeds, and 2 manipulation types), but keeps the abstract manipulations the same. 

We expect models to show seemingly positive effects of preemption for these verbs. This is because in the \removep{} condition, the models can still learn transitive preferences for these verbs from their unaffected transitive cases (by contrast, the target verbs are intransitive, and do not occur with transitive usages, so there is no direct evidence for them). Similarly, in the \keepp{} case, it is likely that a significant amount of non-preemptive usages for these verbs that are removed are transitive. Thus, the models should find the transitive sentences in our test set to be more surprising in the \keepp{} condition than the \removep{} condition.

\subsection{Results and Analysis}

We compute the NLLs (as described in \cref{sec:measure}) for the transitive-sentence subset of our evaluation dataset across all our manipulations. We then analyze our results using a linear mixed-effects regression \citep{bates2015fitting}, using the NLL as our dependent variable, the target verb (\texttt{verb}) and the training set variant ($\mathcal{D}$, \keepp{} vs. \removep{}) and their interaction as fixed effects, and the random seed (\texttt{seed}) and evaluation sentence ($s^-_i$) as random effects. 
This is specified by the following equation:
\begin{align*}
\ell(s_i^-) \sim \texttt{verb} \times \mathcal{D} + (1 \mid s^-_i : \texttt{seed})
\label{eq:lmer-loss}
\end{align*}
We compute this separately for the verb-specific and abstract conditions, and analyze the contrast between \keepp{} and \removep{} across verbs. Insofar as the model shows an effect of preemption, we should expect this contrast to be positive---i.e., the average NLL for \removep{} should be lower than that of \keepp{}. This is because the preemptive evidence is not present in \removep{}, so the unconventional form should no longer be blocked by the model and therefore should be likely.

\begin{figure}[t]
  \includegraphics[width=\columnwidth]{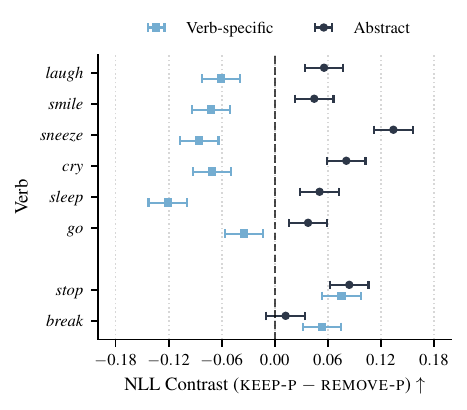}
  \caption{Contrast of estimated marginal means between \variantKeepP{} and \variantRemoveP{} on the linear mixed-effects model, with 95\% confidence intervals, across verb-specific (blue, square) and abstract (black, circle) conditions. Positive values indicate preemption. We additionally include results on two alternating verbs---\textit{break} and \textit{stop}.}
  \label{fig:emm-transitive}
\end{figure}

\Cref{fig:emm-transitive} shows our results across verbs and manipulation conditions. For our target intransitive verbs in the verb-specific condition, instead of seeing a privileged (positive) effect of preemption, we instead see largely negative effects across verbs. That is, the models' NLL on the unconventional, transitive usages for our target verbs is generally \textit{lower} when the hypothesized preemptive evidence is removed, than when it is kept. This is in direct opposition to what is specified by the preemption proposal \citep{goldberg1995constructions, brooks2002does, boyd2011learning, ambridge2015preemption}.

For the abstract condition, however, we see relatively stronger positive effects of preemption than in the verb-specific case. 
That is, our LMs are more affected by the \textit{accumulated} indirect negative evidence from periphrastic causative usages of \textit{all} verbs than by the evidence only for a single verb at a time. This suggests that preemption effects (if any) are more likely to be an abstract effect rather than item-specific, at least from the statistics contained in CHILDES.\footnote{These results hold even if we remove periphrastic causative usages only for all intransitive verbs, leaving those of alternating verbs in the training set. See \cref{sec:appendix-abstract-intrans} for these results.}

On comparing these results to those from performing controlled rearing on \textit{break} and \textit{stop}, we see that the latter indeed show more seemingly positive evidence of verb-specific preemption---per our expectation. But it is important to note that for these verbs, direct evidence for transitive usages still exists in the \removep{} condition. Therefore, even if there are no periphrastic causative usages in this condition, LMs can still pick up on the explicit transitive usages for both verbs (\textit{John broke the cup}, \textit{Mary stopped the car}, etc.).

\subsection{Interim Discussion}
Our verb-specific results from the previous experiment suggest that CHILDES-trained LMs find transitive usages of (generally) intransitive verbs (like \textit{cry} and \textit{smile}) \textit{less}---as opposed to \textit{more}---likely when their periphrastic causative usages are removed. That is, the models do not treat the periphrastic causative as an indirect negative evidence. 

A plausible explanation of why this might be the case comes from earlier controlled rearing results, where models were shown to be sensitive to indirect \textit{positive} evidence. For instance, \citet{misra-mahowald-2024-language} show that knowledge of constructions such as \textit{a beautiful five days} can come from other cases where a plural measure noun phrase (\textit{6 months}) is construed as a singular unit (e.g., \textit{6 months \textbf{is} all I need...}) even in the absence of direct evidence \citep[see also][]{jumelet-etal-2021-language, patil2024filtered, yao2025both}. Therefore, instead of encoding the \textit{competition} between periphrastic causative and transitive constructions, the models might only be encoding their similarity.\footnote{It is important to note that encoding similarity is a precursor to encoding competition---the learner must recognise what structures are similar before concluding that they also compete \citep{suttle2011partial, goldberg2016partial}.} This can arise from alternating verbs such as \textit{tickle, boil, melt, move, grow}, etc. that \textit{can} be used in both constructions. This is perhaps why we see an effect of preemption in the abstract condition, where \removep{} effectively removes evidence that the transitive and periphrastic causative can be related.

We will explore this hypothesis of indirect \textit{positive}---and not \textit{negative}---evidence, by turning to a training dynamics analysis in the next section.

%% file: chapters/06-experiment-moment.tex
\section{Prospective Moments of Indirect Negative (or Positive?) Evidence}
\label{sec:prospective-ine}


In this section, we explore model generalization behavior at an extremely fine-grained level, by testing it at specific time-steps during pretraining. Specifically, we identify `Prospective Moments of Indirect Negative Evidence', which we define as training batches that contain constructions that may serve as the hypothesized negative evidence against overgeneralization. On identifying such cases, we then measure the model's behavior before and after encounter with a batch of interest, to quantify its effect (given encounters with previous batches) on the model's overgeneralization behavior.

\begin{figure}[!t]
    \centering
    \includegraphics[width=\columnwidth]{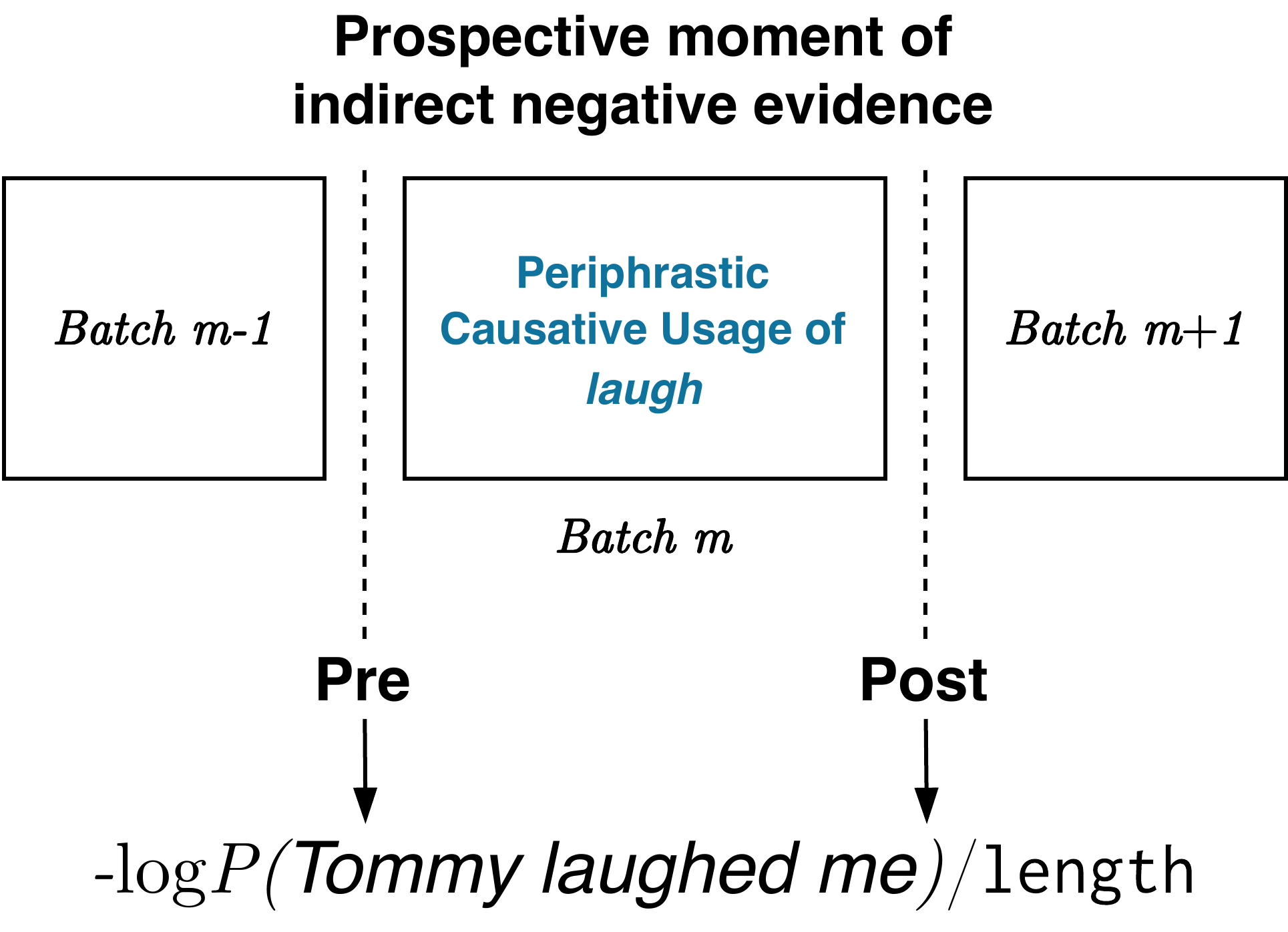}
    \caption{Depiction of a \textit{Prospective Moment of Indirect Negative Evidence}. We identify batches that contain the periphrastic causative usage of a verb (here, \textit{laugh}) and then measure the NLL of an evaluation sentence before (pre) and after (post) encountering the batch.}
    \label{fig:prospective-ine}
\end{figure}

In the context of the previous experiment's findings, we specifically narrow in on batches containing the periphrastic causative usages of verbs. This is because, according to preemption, these were meant to be sources of indirect negative evidence, but our results suggest that the opposite is true. That is, models trained on corpora where this evidence was removed also found transitives to be \textit{less} likely than when this evidence was not removed. This suggests that these moments might in fact be supplying models with indirect \textit{positive} evidence that the two structures are similar.

\begin{figure*}[!t]
    \centering
    \includegraphics[width=\textwidth]{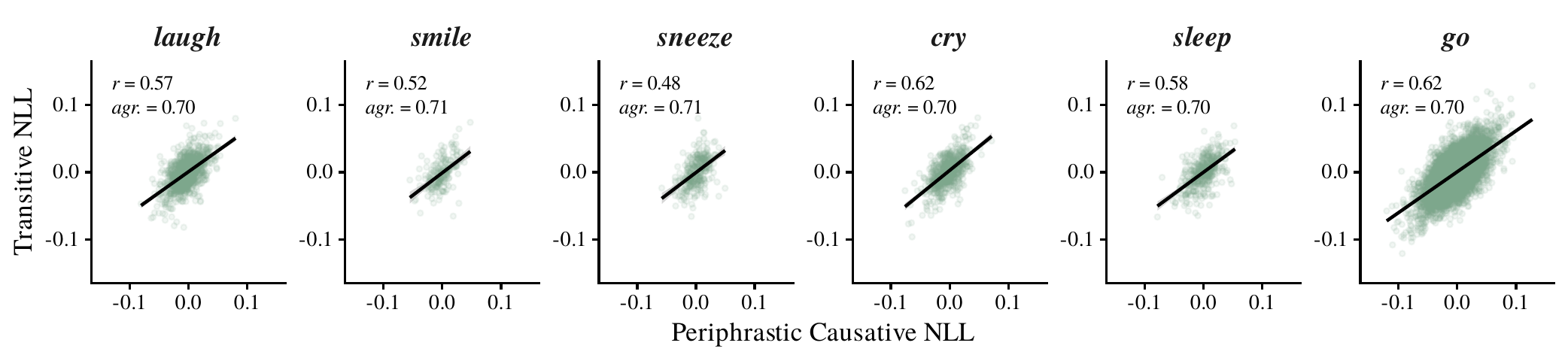}
    \caption{Statistical relationship between transitive and periphrastic causative NLLs, measured at every batch during training that contains a periphrastic causative exposure (indicated by individual points) to the verb of interest.}
    \label{fig:prospective-ine-corr}
\end{figure*}

To shed light on this hypothesis, we measure the correlation between the NLL on our evaluation data's periphrastic causative and transitive usages for each verb, using the models (5 different seeds) trained on the full CHILDES dataset. Specifically, for a batch containing a given verb's usage in the periphrastic-causative construction, we compute the difference between a model's average NLL on an evaluation set before and after encountering that batch. We do this for both periphrastic causative and transitive evaluation sentences. We then measure the Pearson's correlation and the pairwise agreement between the NLL differences for both constructions. Agreement is measured as the percentage of time both values had the same sign. A positive correlation and high agreement would indicate that the losses on both structures change in the same way upon encounters with a periphrastic causative usage during training. For reference, according to preemption, we should instead expect a negative correlation and agreement---the loss on transitives should increase with exposure to a periphrastic causative, whereas loss on periphrastic causatives should decrease.

\Cref{fig:prospective-ine-corr} shows our results. We see generally high correlation and agreement scores across all verbs, corroborating the indirect positive evidence hypothesis. That is, on encountering a periphrastic causative usage of a verb during training, the change in model's behavior on both periphrastic causatives \textit{and} transitives are positively correlated.

%% file: chapters/99-closing.tex

\section{Implications and Speculations}

We find that at least at the verb-specific level, LMs trained on CHILDES treat the periphrastic causative construction as indirect evidence \textit{for}, and not \textit{against}, transitive usages of intransitive verbs. Below we discuss the broader implications and speculations of these findings: 

\subsection{Semantic cues against overgeneralization} Preemption as a mechanism attributes a more special role to the communicative intent of contexts in which the preemptive evidence is used as compared to entrenchment. Models' failure to show effects of preemption in the verb-specific condition shows that the language-only cues in CHILDES are not sufficient to give rise to this sensitivity in models. But this also raises the question about whether providing explicit semantic evidence of events could facilitate preemption in LMs. This could come from environmental annotations---cues to what is going on in the child's environment at the time of the utterance---that are made available in CHILDES (but were not used in our work, though see \citet{wu2025mechanistic}, who did use them). This could also be induced through more explicit means, like in \citet{yedetore-kim-2024-semantic}, who showed evidence of hierarchical generalization in transformers when semantic signals (in the form of semantic parses) were available during training.

\subsection{Morphological origins of preemption}
While we investigated preemption for argument structure constructions in this work, the phenomenon traces its origins to morphological `blocking' \citep{aronoff1976word, kiparsky1973elsewhere, clark1987principle}. For example, \textit{children} preempts \textit{childs} or \textit{went} preempts \textit{goed}. Availability of such a phenomenon during learning could be the key to preemption. Insofar as this is true, a hypothesis for why we might not see strong effects of preemption in our LMs is that tokens like \textit{childs} and \textit{goed} are simply not represented in neither our BPE nor our lemmatized variants, and so the model is never preempted in the morphological sense. This motivates serious consideration of morphologically-informed tokenizers in the development of LMs.

\subsection{Preemption from abstract cues} 
While we do not observe preemption in our verb-specific experiments, we do see evidence of more abstract signals to preemption via exposures from other verbs. 
In the context of human language acquisition, preemption has always been treated as a verb-specific phenomenon \citep{goldberg2005constructions, ambridge2015preemption, ambridge-etal-2018-effects, bidgoodVerbArgumentStructure2021}. Our results suggest the possibility of preemptive signals coming from more accumulated sources of indirect evidence---e.g., transitive usages of \textit{laugh} could come from periphrastic causative usages of \textit{laugh}, combined with those from other verbs (like \textit{giggle}, \textit{smile}, etc.). This motivates \textit{novel} human experiments, perhaps via artificial language learning paradigms \citep[e.g.,][]{samara2025learners}, to test if this is a viable route for human learners. Importantly, this does not necessarily invalidate any potential finding of preemption in humans, which have been concluded from using corpus evidence in the past \citep{boyd2011learning, ambridge2015preemption, ambridge-etal-2018-effects, bidgoodVerbArgumentStructure2021}. However, it does raise the question whether such estimates (which are done at the verb level) might by strengthened if the effect of other verbs' usages is also considered. 

\section{Conclusion}
Overall, our work joins several others in understanding how the input to a general purpose statistical learner like a transformer LM shapes its generalization behavior. In particular, it showcases how subtly different proposals for indirect negative evidence can be disentangled, especially when doing so is impossible for human learners \citep{ambridge2015preemption}. In closing, this leaves us optimistic about the potential of LMs to contribute in-principle insights about how statistical learning might unfold in language learners \citep{contreras2023large, futrell2025linguistics, misra2024generating}. 

\section*{Limitations}

\paragraph{Single Phenomenon}
Due to the number of pretraining experiments, we focus on a single phenomenon, although it is among the quintessential examples of this debate which has lasted around 40 years \citep{bowerman1988no, akhtar1997young, brooks1999children, brooks2002does, theakston2004role, ambridge2008effect, ambridge2015preemption, ambridge-etal-2018-effects, bidgoodVerbArgumentStructure2021}. It is possible that models show different generalization behaviors on different constructions. Our methodology, and the methodology of controlled rearing in general can be used to adjudicate between preemption and entrenchment across several different constructions in the future. 

\paragraph{Generalization of our conclusion across models}
Our analyses have only looked at one instance of the entire class of transformer LMs that can be trained on CHILDES, finding lack of evidence for preemption at the level of individual verb exposures. However, this does not necessarily mean that it cannot arise in transformers or neural networks in general. In fact, there is a connectionist model that does show effects of both preemption and entrenchment \citep{ambridge2016connectionist}---however its architecture explicitly bakes in hand-coded semantic features which have to be supplied by the researcher for a given input. That is, this model is at a different level of analysis and theoretical commitment than ours, and cannot learn from naturalistic data. Furthermore, it would also be worthwhile to understand preemption from a theory of computation perspective, in order to provably analyze preemption in classes of transformers (or other types of neural network models). Finally, it is possible for these effects to be stronger with larger datasets, e.g., corpora from the BabyLM challenge \citep{choshen2026babylm}, or larger models trained on such larger datasets.


\paragraph{Operationalizing preemptive signals}
We have assumed that all instances of the competing construction constitute cases where either construction could have been used. But that need not be true. More generally, there is a need for a methodology to detect the possibility of competition vs. non-competition from raw data. Additionally, it might be too narrow of an assumption to treat only near-synonymous constructions as preemptive \citep{ambridge2015preemption}. For instance, in a scenario where something (a joke) makes someone laugh (John), it might be perfectly ok to simply use the intransitive, \textit{John laughed}. In this sense, the intransitive could have preempted the transitive usage. Therefore, future work could address this by running controlled rearing on specific cases where constructions necessarily compete vs. do not. Finally, what counts as a ``semantic context'' can itself differ. For instance, in \citet{misra2024generating}, the features of the constructional exposure itself ended up showing an effect of preemption. Their case focused on the dative alternation \citep[DO: \textit{John gave me the book} vs. PO: \textit{John gave the book to me};][]{levin-1993-english} and found that the information structure of the theme (the book) and recipient (me) predicted the degree of preemption. Future work can hopefully consolidate the many different ways preemption can be cued to learners.




\section*{Acknowledgments}

We are grateful to Adele Goldberg, Caroline Rowland, and Qing Yao for valuable comments and advice on earlier versions of our experiments. We also acknowledge Reviewer zAa3 for their thoughtful review during the ARR May Cycle---their comments resulted in our control experiments. KM was supported by a Donald D. Harrington Faculty Fellowship at UT Austin for the 2025/26 academic year. FS was supported by a Canada CIFAR AI Chair award.

%% file: chapters/a1-technical.tex

\section{Details for Training}

Additional details for training hyperparameters are included in \Cref{tab:training-hyperparameters}.
We pack the consecutive utterances in a single CHILDES file into sequences of 1024 tokens, and the \texttt{<s>} token is used as the beginning of sentence and also the separator between two sentences.
We use the GPT-2 implementation provided by Huggingface Transformers \citep{wolf-etal-2020-transformers} and a custom PyTorch Lightning training loop for training our LMs, and a single training run requires around 1.5 hours on 1 NVIDIA L40s GPU.

\begin{table}
\centering
\begin{tabular}{ll}
\toprule
\textbf{Hyperparameter} & \textbf{Value} \\
\midrule
Max Sequence Length & 1024 \\
Hidden Dim Size & 768 \\
Number of Layers & 12 \\
Number of Attention Heads & 12 \\
Token per Batch & 8192 \\
Optimizer & AdamW \\
Learning Rate & 1e-5 \\
Learning Rate Scheduler & Linear \\
Warmup Steps & 1500 \\
Weight Decay & 0 \\
$\beta_1$ & 0.9 \\
$\beta_2$ & 0.95 \\
Gradient Clip Norm & 1.0 \\
\bottomrule
\end{tabular}
\caption{Hyperparameters used for training our LMs.}
\label{tab:training-hyperparameters}
\end{table}

\section{Identification of Periphrastic Causative Constructions}
\label[appendix]{sec:causative-identification}

\begin{figure*}[!t]
    \centering
    \includegraphics[width=\linewidth]{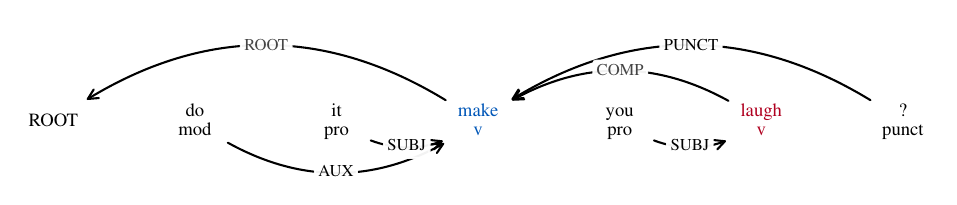}
    \caption{An example of the dependency parse trees on the CHILDES dataset. This is Eng-NA/Bloom/Peter/020303, sentence 1028.}
    \label{fig:example-parse}
\end{figure*}

We use PyChildes \citep{ma-etal-2025-pychildes} to parse the raw CHILDES files of Eng-NA and Eng-UK, and then extract morphological and dependency grammar annotations.
These two corpora of CHILDES are open to public access.
An example of the dependency parse trees is shown in \Cref{fig:example-parse}.
Then we use the following heuristics to identify periphrastic causative constructions in the CHILDES dataset.

For \textit{let} and \textit{make}, we identify them as matrix verbs in a periphrastic causative construction if they satisfy the following conditions:
\begin{itemize}[topsep=2pt, itemsep=-3pt, parsep=0.5em, leftmargin=*]
\item Its complement (\texttt{COMP}) child is a verb not in a to-infinitive (e.g. \textit{make something to eat}).
\item Either it has an object (\texttt{OBJ}) child (e.g. \textit{make me laugh}), or its complement verb has a subject (\texttt{SUBJ}) child (e.g. \textit{make him see it}).
\end{itemize}

Additionally, for \textit{let}, we require it to have a subject (\texttt{SUBJ}) within its predicate to exclude hortatives like \textit{let's go}.
This is done by searching through its ancestor chain formed by complement (\texttt{COMP}), coordination (\texttt{COORD}), and conjunction (\texttt{CONJ}) arcs.

For \textit{get}, \textit{cause}, and \textit{force}, we use the same criteria except that the complement verb must be in a to-infinitive (e.g. \textit{get him to sleep}).

Finally, for \textit{have}, we use a similar method as \textit{let} and \textit{make}, and then use an external LLM, Gemini 3 Flash,\footnote{\url{https://ai.google.dev/gemini-api/docs/models/gemini-3-flash-preview}} to filter out false positives. We use the following prompt:
\begin{Verbatim}[breaklines=true, breakanywhere=true, fontsize=\small]
In the following sentences, some contain a causative *have* and some do not.
"May I have more juice please ?" (False)
"I have finished my juice ." (False)
"I had the waiter bring more juice ." (True)
"I had my juice refilled by the waiter ." (True)

Analyze the following sentence and determine whether it contains a causative *have* or not. Answer with True if it does and False if it does not.
"{sentence}"
\end{Verbatim}

We include the statistics of some more verbs that we identify in their usage in CHILDES as having periphrastic causative constructions in \Cref{tab:extra-verb-counts}.

\begin{table}[!t]
\centering
\resizebox{0.65\columnwidth}{!}{%
\input{assets/tables/extra_verb_count.tex}
}
\caption{Additional verb counts and periphrastic causative usage counts in the CHILDES dataset. This table is not exaustive.}
\label{tab:extra-verb-counts}
\end{table}

\section{Details for Evaluation Data Synthesis}

We use Qwen 3.5 27B \cite{qwen3.5} to synthesize evaluation sentences. Our system prompt is as follows:
\begin{Verbatim}[breaklines=true, breakanywhere=true, fontsize=\small]
You are a child literature author tasked with choosing objects that fit naturally into simple, easy-to-understand scenes for young children's literature.
\end{Verbatim}

The main prompt is as follows, where \texttt{previous\_objects} contains samples of previously used objects for this verb.
For \texttt{subject} we sample from 266 names and pronouns, and for \texttt{object\_candidates} we sample from names, pronouns and additionally 499 nouns like \textit{baby}, \textit{mommy}, \textit{toy}.
The language model is prompted to choose an object from 25 randomly sampled candidates that fits most naturally with the subject and verb.
The chosen object, along with the subject and verb, are then used to form the periphrastic causative sentence in the evaluation set.
The ungrammatical transitive counterpart is then generated by mutating the periphrastic causative sentence.

\begin{Verbatim}[breaklines=true, breakanywhere=true, fontsize=\small]
### Instructions
Choose the object from the candidates below that best fits naturally with the subject and verb, \
forming a plausible scene where {usage_description}, as in the sentence "{usage_example}".
If the chosen object is a common noun, prefix it with the article "the" (e.g. "plant" -> "the plant"). \
If the chosen object is a proper name (e.g. a person's name), do not add an article (e.g. "Tom" stays "Tom").
Where the object is the one who will perform the action (not receive it), prefer a person or animal over an inanimate noun.
If none of the candidates are suitable, answer with "none".

### Output Format
Answer with only the chosen object, exactly as written in the candidate list but with the article \
added if required, or "none".

### Input
- Subject: {subject}
- Verb: {verb}
- Object Candidates: {object_candidates}{already_selected_line}
- Already selected for this verb (prefer a different one): {previous_objects}"
\end{Verbatim}

Additionally, we lemmatize the evaluation set with Spacy, using model \verb|en_core_web_trf| version 3.8.0.

%% file: assets/tables/extra_verb_count.tex
\begin{tabular}{lrr}
\toprule
\textbf{Verb} & \textbf{PC Usage} & \textbf{Verb Usage} \\
\midrule
\textit{be} & 272 & 4150 \\
\textit{break} & 11 & 10958 \\ 
\textit{change} & 37 & 3007 \\
\textit{close} & 18 & 4822 \\
\textit{come} & 253 & 88850 \\
\textit{cook} & 15 & 3538 \\
\textit{cough} & 22 & 443 \\
\textit{cry} & 49 & 6166 \\
\textit{do} & 299 & 153550 \\
\textit{drink} & 19 & 6209 \\
\textit{eat} & 113 & 44783 \\
\textit{fall} & 142 & 15746 \\
\textit{feel} & 240 & 6237 \\
\textit{get} & 260 & 178352 \\
\textit{go} & 1295 & 316515 \\
\textit{grow} & 49 & 2871 \\
\textit{have} & 439 & 147278 \\
\textit{jump} & 65 & 5450 \\
\textit{know} & 106 & 92370 \\
\textit{laugh} & 99 & 1781 \\
\textit{like} & 93 & 54171 \\
\textit{look} & 196 & 70121 \\
\textit{make} & 163 & 53323 \\
\textit{move} & 43 & 7448 \\
\textit{open} & 26 & 10193 \\
\textit{play} & 198 & 37279 \\
\textit{put} & 173 & 99050 \\
\textit{roll} & 16 & 3555 \\
\textit{run} & 53 & 9114 \\
\textit{see} & 275 & 101864 \\
\textit{sing} & 24 & 7331 \\
\textit{sit} & 150 & 27679 \\
\textit{sleep} & 33 & 8614 \\
\textit{smile} & 15 & 724 \\
\textit{sneeze} & 31 & 461 \\
\textit{stand} & 132 & 7124 \\
\textit{stay} & 72 & 8706 \\
\textit{stop} & 42 & 5826 \\
\textit{swim} & 12 & 3901 \\
\textit{talk} & 58 & 10857 \\
\textit{think} & 94 & 78822 \\
\textit{turn} & 23 & 12212 \\
\bottomrule
\end{tabular}

%% file: chapters/a2-additional-results.tex

\section{Final Checkpoint Results for the Precondition Experiment}
\label{app:final-ckpt-exp1}

\begin{figure}
    \centering
    \includegraphics[width=\columnwidth]{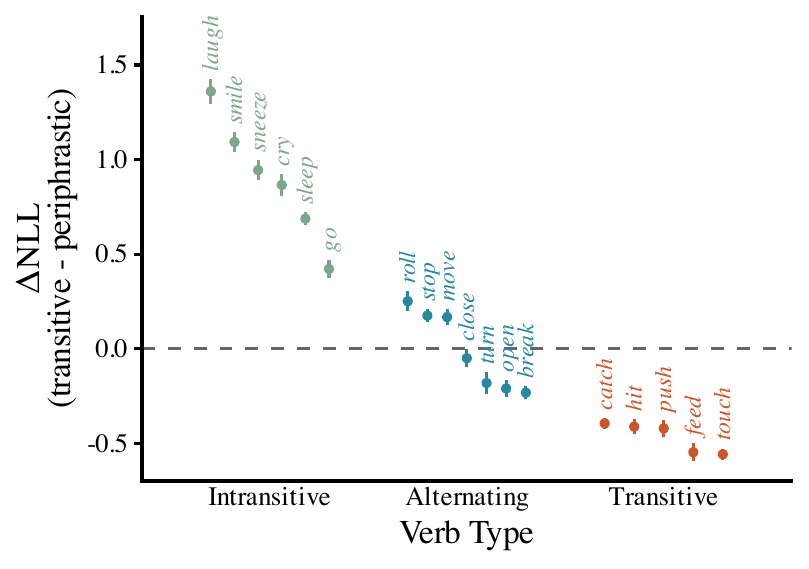}
    \caption{Final Checkpoint $\Delta$NLLs across 5 seeds and 100 sentence pairs, per verb, across verb types.}
    \label{fig:final-ckpt-exp1}
\end{figure}

\Cref{fig:final-ckpt-exp1} shows the $\Delta$NLLs for the different verbs in the precondition experiment (\Cref{sec:precondition}). We again see clean separation of model preferneces across different verb classes (intransitive, alternating, transitive).

\section{Additional Results for Abstract Effect from Intransitive Verbs}
\label[appendix]{sec:appendix-abstract-intrans}

\begin{figure}[t]
  \includegraphics[width=\columnwidth]{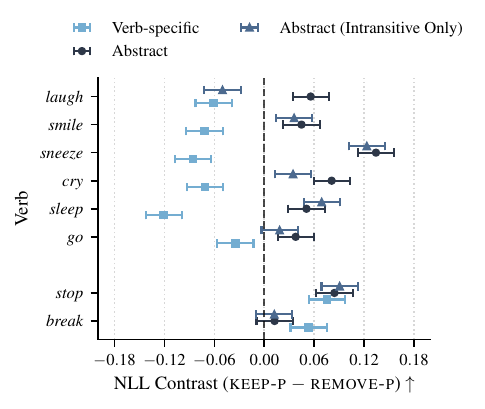}
  \caption{Contrast of estimated marginal means between \variantKeepP{} and \variantRemoveP{} on the linear mixed-effects model, with 95\% confidence intervals, across verb-specific (blue, square), abstract intransitive only (dark blue, triangle), and abstract (black, circle) conditions. Positive values indicate preemption.}
  \label{fig:emm-transitive-intrans-only}
\end{figure}

In addition to \Cref{subsec:preemption-manipulations}, we test a manipulation condition in between verb-specific and abstract, where we remove all periphrastic causative usages of all 174 intransitive verbs we identified in the CHILDES dataset. 
Periphrastic causative usages of alternating verbs are kept intact.
This results in removing 4,675 usages, 62.7\% of all periphrastic causative usages in the CHILDES dataset.
In the total abstract setting, \removep{} models do not observe any periphrastic causative construction during training, while in the intransitive only setting, \removep{} models still observe periphrastic causative constructions of alternating verbs.
This allows intransitive only \removep{} models to establish the connection between periphrastic causative and transitive constructions.
Results are shown in \Cref{fig:emm-transitive-intrans-only}.
We see that the abstract effect from the intransitive verbs to intransitive target verbs are close to the total abstract effect, usually slightly weaker, at the exception of \textit{laugh} (which shows a negative effect closer to the verb-specific effect) and \textit{sleep} (which has an insignificantly stronger positive effect than the total abstract effect).
The observed positive effect in preemption indicates that preemptive signals from other intransitive verbs---the verbs of same syntactic category---contribute a substantial portion of the total indirect evidence for the target verbs' retreatment from overgeneralization.

\section{Additional Results for Training Trajectories}

\begin{figure*}[!t]
    \centering
    \includegraphics[width=\textwidth]{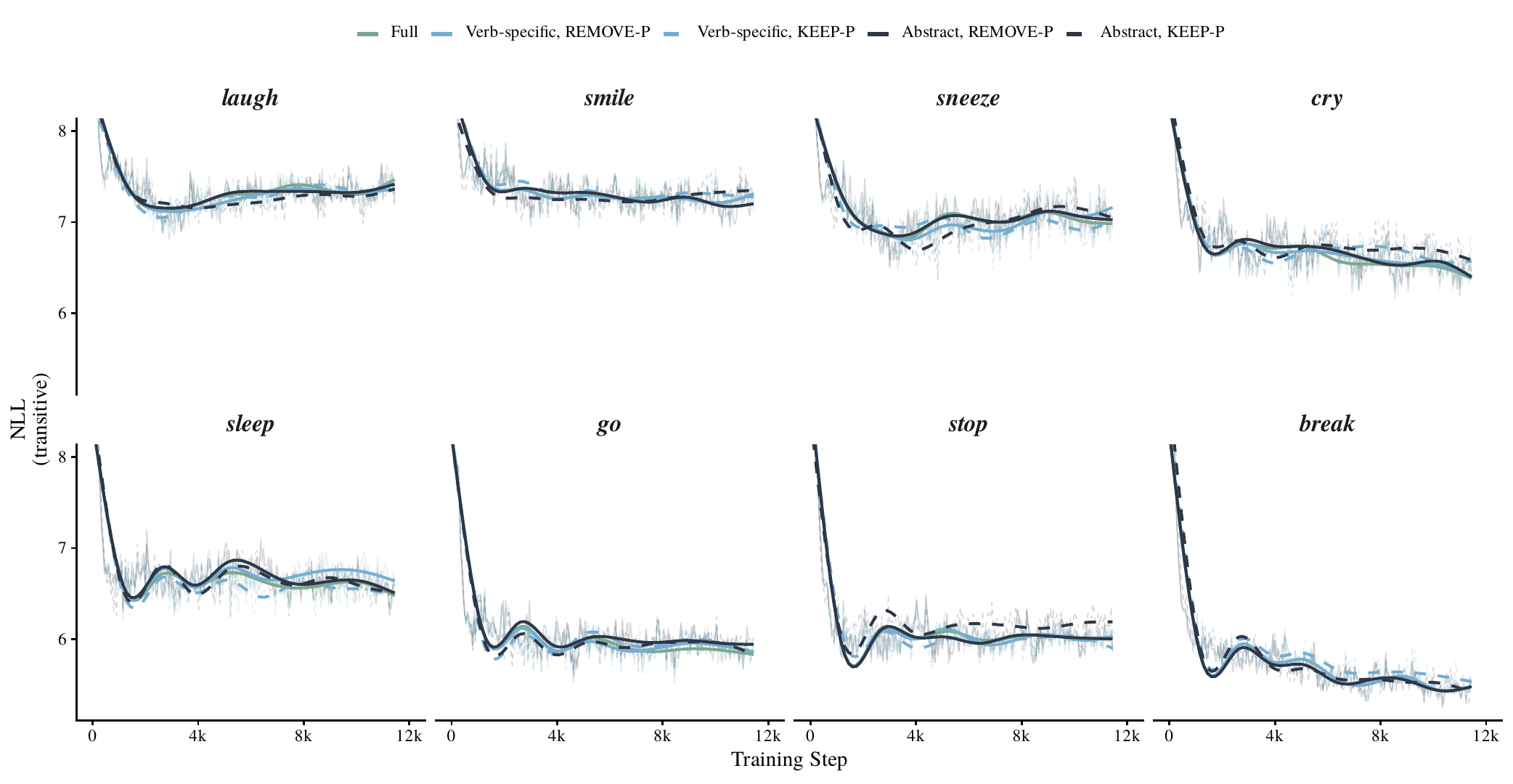}
    \caption{Difference in NLL of transitive and periphrastic causative sentence pairs for 6 target intransitive verbs and 2 alternating verb controls at every 50 steps during controlled rearing training. Each line represents a different training run using the same seed. \removep{} models are shown in solid lines, while \keepp{} models are shown in dashed lines.}
    \label{fig:precondition-for-rearing}
\end{figure*}

We further plots the training trajectory of our controlled rearing conditions in \Cref{fig:precondition-for-rearing}, similar to \Cref{fig:precondition-results}.
As discussed in \Cref{sec:measure}, we show the NLL of overgeneralized transitive form only, instead of the $\Delta$NLL between transitive and periphrastic causative forms.
Due to the aggregation over different evaluation sentences, the NLLs between different controlled rearing conditions shed the consistent pairwise differences we shown in \Cref{fig:emm-transitive},
and the smoothing of curves makes it ineligble to compare the final NLL we used in the main experiments.
However, results here indicate that our models largely converging under 3 epoches of training without significant sign of overfitting.
And they also indicate that the overgeneralization tendency can be volatile during training.
We leave more systematic analysis of training trajectories to future work.

\section{Additional Results Without Lemmatization}
\label[appendix]{sec:appendix-bpe}

\begin{figure}[h]
  \includegraphics[width=\columnwidth]{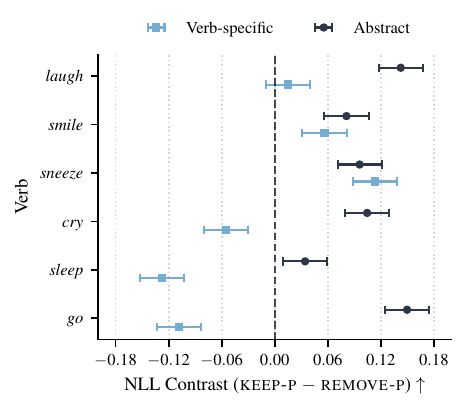}
  \caption{Contrast of estimated marginal means between \variantKeepP{} and \variantRemoveP{} LM variants trained and evaluated on BPE-tokenized versions of datasets.}
  \label{fig:emm-transitive-bpe}
\end{figure}

Due to the nature of our training data and LM size, the commonly used Byte-Pair Encoding (BPE) subword tokenization can introduce extra burden to the training process.
For example, the model will have to learn the connection between different word forms of \textit{make} (e.g., \textit{made}, \textit{making}, etc.), which will be tokenized into different subword sequences.
Nevertheless, we repeat our main analyses under a BPE subword tokenization setting.
We first train a 8,192-token BPE tokenizer on the CHILDES dataset, following \citet{huebner-etal-2021-babyberta, misra2024generating}.
Then in both training and evaluation data pipelines we remove the lemmatization components, and swap the original word-level tokenizer with this BPE tokenizer.
We then repeat the verb-specific and abstract condition in \cref{subsec:preemption-manipulations} for 6 verbs and train 70 models (5 seeds $\times$ 6 verbs $\times$ 2 manipulation types for verb-specific, 5 seeds $\times$ 2 manipulation types for abstract),
and show results in \cref{fig:emm-transitive-bpe}.
Under abstract level manipulation, the effects are stronger than the verb-specific case, except for \textit{sneeze} and \textit{smile}, whose verb-specific effects also flipped from negative to positive.
The results are largely consistent with those using lemmatization.

%% file: chapters/a3-guo.tex
\section{A Reassessment of \citet{guo2026do}}
\label{sec:rebuttal-guo}

Our work is most closely related to that of \citet{guo2026do}, who measured preemption and entrenchment effects across three different linguistic phenomena (including transitive overgeneralizations), and across multiple off-the-shelf LMs. They operationalize overgeneralization as the difference in LM surprisal (negative log-probability) of the unconventional (overgeneralized) sentence vs. that of the conventional sentence. In the case of transitive overgeneralizations, their method measures difference in the surprisal of sentences like \textit{*Tommy laughed me} and those like \textit{Tommy made me laugh}. Using this measure, they find evidence for preemption over entrenchment in LMs. They do so by first showing high correlation of the difference in surprisals with preemptive usages of verbs, even when controlling for the non-preemptive usages. Then, they fine-tune an off-the-shelf GPT-2 model \citep{radford-etal-2019-gpt2} on 5000 instances of conventional vs. unconventional usages, finding overwhelming effects of preemption over non-preemption.

On further scrutiny of \citet{guo2026do}'s methods, we find their results to paint an illusory picture about preemption in LMs. We highlight two concerns. First, they use fine-tuning as their method to claim causal evidence of preemption. However, an LM might have already formed its preferences of avoiding the target overgeneralizations during pre-training (as shown by their results). 
So, the results from these experiments do not sufficiently shed light on how the dispreference of overgeneralization was learned in the first place---this will require training data manipulations. Second, and more importantly, their method explicitly includes the preemptive construction as part of the overgeneralization measure (e.g., \textit{periphrastic causative} for \textit{transitive}). That is, the conventional form in their measure is exactly the same form that is hypothesized to preempt the unconventional form. Therefore, it is unsurprising that an effect of preemption is observed on fine-tuning LMs on the conventional forms---change in the measure is directly being manipulated in their fine-tuning experiment. So any effect of preemption might just be driven by the increased likelihood of the conventional form rather than the unlikelihood of the unconventional form. 

Disentangling preemption from entrenchment will require direct manipulation of the training data, and a more precise measure of overgeneralization that focuses exclusively on the unconventional form. Our methods in \cref{sec:method} aim to embody this conclusions directly.




%% file: chapters/a4-alt.tex
\section{Alternate Methods of Measuring Overgeneralization Effects}
\label{sec:guo-results}

\begin{figure}[!t]
\includegraphics[width=\columnwidth]{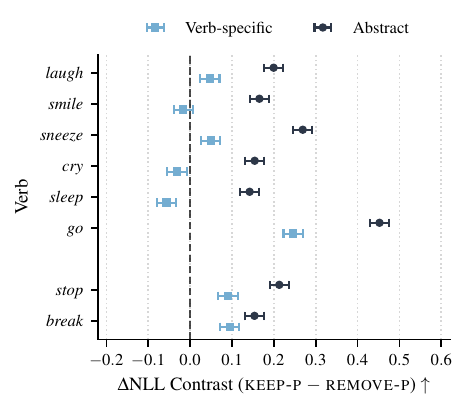}
\caption{Contrast between \variantKeepP{} and \variantRemoveP{} on a similar linear mixed-effects model as in \cref{fig:emm-transitive}, but with the NLL difference between the transitive and periphrastic causative sentences as the dependent variable. This is the result we would get had we used \citet{guo2026do}'s method.}
\label{fig:emm-nll-difference}
\end{figure}

In this section, we compare our results to those if we had used \citet{guo2026do}'s method of treating the difference in NLL of transitive and periphrastic causative usages as our measure of overgeneralization effects. We replicate the above analysis using this measure (computing NLLs of both subsets of our evaluation data), and show results in \cref{fig:emm-nll-difference}. We see that on \textit{go}, \textit{laugh}, \textit{sneeze} there is seemingly positive effect of preemption.
But as our previous \Cref{fig:emm-transitive} reveals that the NLL of transitive is actually lower in \keepp{} than in \removep{}, we can conclude that the positive effect observed here is driven by the NLL of the periphrastic causative loss being lower in \keepp{}.

